\documentclass[conference]{IEEEtran}
\IEEEoverridecommandlockouts
\usepackage{cite}
\usepackage{amsmath,amssymb,amsfonts}
\usepackage{graphicx}
\usepackage{textcomp}
\usepackage{xcolor}
\usepackage[ruled,vlined]{algorithm2e}
\SetKwProg{Fn}{Function}{:}{}

\def\BibTeX{{\rm B\kern-.05em{\sc i\kern-.025em b}\kern-.08em
    T\kern-.1667em\lower.7ex\hbox{E}\kern-.125emX}}
\begin{document}


\title{Online Planning for Sparse Ground Target Search from a High-Altitude UAV under Partial Observability}

\author{
    \IEEEauthorblockN{Ashik E Rasul}
    \IEEEauthorblockA{\textit{Department of Mechanical and Nuclear Engineering} \\
    \textit{Tennessee Technological University}\\
    Cookeville, USA \\
    ashik.rasul@outlook.com}
    \and
    \IEEEauthorblockN{Hyung-Jin Yoon}
    \IEEEauthorblockA{\textit{Department of Mechanical and Nuclear Engineering} \\
    \textit{Tennessee Technological University}\\
    Cookeville, USA \\
    hyoon@tntech.edu}
}


\maketitle

\begin{abstract}

Unmanned aerial vehicles (UAVs) searching for ground targets from a high altitude face a unique challenge, particularly when the target is already within the field of view but effectively unobservable because of its small apparent scale. Standard object detectors often underperform in such scenarios because of resolution downscaling and limited context. In contrast, active object search frameworks address this challenge by directing the agent to a suitable pose to gather richer visual information. However, flight regulations in urban airspace often restrict such physical movements for UAVs. As an alternative active sensing approach, the UAV can leverage the pan-tilt-zoom (PTZ) mechanism of the onboard camera to dynamically adjust its field of view and sequentially gather enhanced visual information from specific regions of interest. Once the candidate locations of the target are identified, it can deploy more expensive object detection schemes, such as an ensemble of multiple models, to get better reasoning at a fixed scale. In this work, we formulate the sequential exploration with PTZ operation as a partially observable Markov decision process (POMDP), in which the agent maintains a belief state over the target's true location. To solve the POMDP, we deploy partially observable Monte Carlo planning (POMCP), where we condition the sensing reliability on target object scale and deploy selective ensemble detection as an additional reasoning step.

We validate our methodology with experiments in a photorealistic simulator under different environmental conditions and vehicle states, showing detection of ground targets at variable scales with significantly fewer steps and minimal dependence on sensor resolution compared to baseline methods

\end{abstract}


\section{Introduction}
Humans are particularly good at searching for relevant information in a cluttered environment through a spatial attention mechanism~\cite{posnerAttentionDetectionSignals,eriksenVisualAttentionField1986}. This capability allows us to detect small objects at a distance, even if they occupy a tiny area within our field of view (FOV). Emulating such adaptive perception capability is crucial for efficient object search and has been extensively studied in the computer vision community as human eye movement models~\cite{butkoIPOMDPInfomaxModel2008,butko2009optimal}. However, these models assume that the object of interest is always present in the input image, that the image already contains sufficient features of the target object, and that the sensor has a high detection probability with negligible noise in the operating range. 

\begin{figure}
    \includegraphics{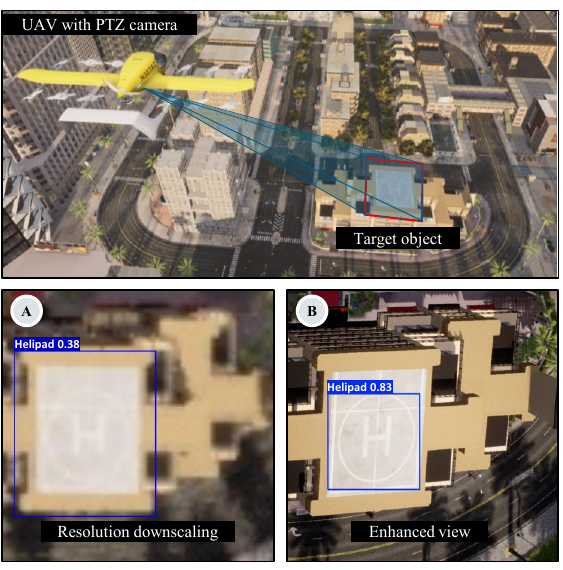}
    \caption{(Top) UAV search scenario for a sparse target from the cruise altitude; (bottom) (a) incoming image downscaling effect. The existing image fails to reveal distinct features for detection after slicing; (b) input image enhancement with camera PTZ operation.}
    \label{fig:UAV_carla}
\end{figure}

\begin{figure*}[t]
    \centering
    \includegraphics[width=\textwidth]{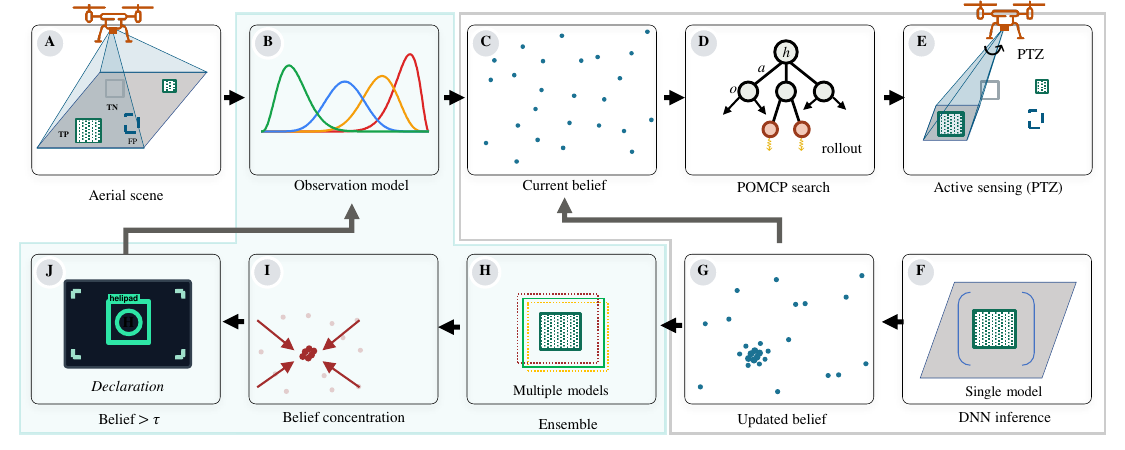}
    \caption{POMCP-PTZ methodology: (a) Aerial search scenario from high altitudes, markers showing TP (True positive), TN(True negative), FP(False positive) detections, (b) Observation model with prior assignment, (C-G) POMCP planning for proposing best PTZ actions at current belief and updating the belief as unweighted particle representation, (H) Prediction with ensemble for increased accuracy, (I-J) Declaration of target while belief crossing the threshold. (H,I,J,B) Observation model formulation and selective ensemble as an extension to regular POMCP.}
    \label{fig:ptz_framework}
\end{figure*}

In a real-world object search scenario, most images collected by a mobile robot may not contain the target object, the target object may be occluded or may not reveal sufficient features, and the detector itself may be noisy. There has been significant advancement in object search in indoor environments by the robotics community, where these challenges are handled by physically moving the agent to a suitable pose to gather enhanced visual information~\cite{xiao2019online, zheng2021multi}. These works assume the observation uncertainty mostly arises from object occlusion or hidden features for the object's relative pose, as the target object scale variation within the FOV is often negligible in indoor settings. For unmanned aerial vehicles (UAVs), searching for a sparse ground target from high altitude, such physical movement is not feasible due to flight regulations in urban airspace. Moreover, even if the target is visible within the FOV, it may be tiny, and the input image resolution may not be enough to reveal sufficient features. In this scenario, standard object detectors often struggle to produce reliable detections. Computer vision methods for tiny object detection also suffer in this case, as a higher-resolution version of the input image may not be available, and approaches such as slicing~\cite{akyon2022slicing}, that attempts to enhance the size of the target by exhaustive slicing of the whole image with overlaps, or coarse-to-fine~\cite{huang2022ufpmp} detection, that runs a coarse detector to find the candidate regions and runs a fine detector on the candidate regions for finding the target, leads to significant resolution downscaling that produces uncalibrated confidence scores and a number of false positive detections. Though the detection accuracy can be further enhanced with more expensive ensemble methods~\cite{solovyev2019weighted}, which run inference with multiple models trained on a subset of data (bagging~\cite{breiman1996bagging}) or with random initialization~\cite{lakshminarayanan2017simple}, they can still struggle under scale and resolution variation of the target image.

The UAV can leverage the pan-tilt-zoom (PTZ) mechanism before capturing the images to selectively enhance the size, clarity, and position of the object of interest, while maintaining a fixed input resolution, as illustrated in Figure~\ref{fig:UAV_carla}. However, such selective enhancement can be expensive, and any exhaustive exploration may be limited by on-board computational resources and time-sensitive mission objectives. Conventional autonomous PTZ camera control follows a preloaded policy, often learned offline, that assumes complete environmental observability and focuses primarily on target tracking rather than ensuring detection reliability~\cite{sandha2023eagle, fang2025automatic}. For the sparse target search scenario, these assumptions do not hold, as the UAV cannot access the full information about the environment simultaneously and can only receive noisy observations with variable uncertainty conditioned on the PTZ settings, environmental conditions, and the target's relative pose. In other words, the environment is partially observable, and the UAV must maintain a belief state, which is an internal estimate of its surroundings, to effectively choose the next sequence of actions. At every time step, the UAV has to trade off between wide-scan exploration for quick area coverage to receive noisy observations and precise-scan exploration for slower area coverage to receive more reliable observations.



In this work, we formulate the sparse target search problem from a high altitude using camera PTZ functionality as a partially observable Markov decision process (POMDP)~\cite{kaelblingPlanningActingPartially1998}. In this setting, the control policy operates directly on the belief space to sequentially actuate the PTZ operation of the camera to optimize the long-term objective of finding the target with minimum steps. The belief gets updated as the UAV receives new observations. Unlike conventional approaches that assume complete knowledge of the environment (i.e., complete observability of the environment), assuming near-perfect sensors or fixed uncertainty across actions, we explicitly model the observation uncertainty conditioned on different object appearances within the FOV.
To solve the POMDP, we deploy Partially Observable Monte Carlo Planning (POMCP)~\cite{silver2010monte}. To address the variation of the observation model in different deployment environments and UAV states, we formulate the detection reliability with a Beta-Bernoulli conjugate model that allows us to incorporate prior knowledge about the observation uncertainty as well as update the model iteratively as we gather more experiences from the deployment scenarios. Additionally, we selectively deploy an ensemble of DNNs at a fixed scale, conditioned on belief concentration, to improve prediction and enable successful target search. 
Our proposed methodology is illustrated in Figure~\ref{fig:ptz_framework}.

To evaluate our approach, we use a simulation environment~\cite{bansal2025airtaxisim} built in Unreal Engine that models realistic aerial scenarios, including cluttered urban landscapes, varying altitudes, camera attributes (Field of view, focal length, and aperture control), and dynamic lighting. This environment enables thorough testing of the proposed framework, allowing us to study the effect of uncertainty, limited observability, and PTZ camera control against the baselines. Our key contributions are: 

\begin{itemize}
    \item We formulate the PTZ camera operation, by a high-altitude UAV searching for a sparse ground target, as a POMDP. 
    
    \item We formulate the observation model uncertainty as a Beta-Bernoulli conjugate model, conditioned on the object appearance, with a prior assignment.


    \item We deploy POMCP to solve the POMDP that sequentially selects the informative actions for environmental exploration. Within the POMCP implementation, we integrate our defined observation model and selectively deploy ensemble detection to enhance reasoning when belief concentration is already high.

    \item Finally, we validate our method in a photo-realistic simulation environment to show the confident discovery of the targets with significantly fewer steps, with minimal dependency on camera resolution, compared to traditional tiny-object detection methods.

\end{itemize}

\section{Background}


POMDPs~\cite{kaelblingPlanningActingPartially1998} provide a compact representation for sequential decision making in real-world scenarios when the environment is not fully observable and provide a principled way to model sensor uncertainty. It is is defined by the tuple $\langle \mathcal{S}, \mathcal{A}, \mathcal{O}, T, \mathcal{Z},R, \gamma\rangle$. At any time step, 
the agent takes an action $a\in\mathcal{A}$ that causes the environment state to transition from $s\in \mathcal{S}$ to $s' \in \mathcal{S}$ following the state transition model $T$. The agent receives an observation $o\in \mathcal{O}$, with observation uncertainty modeled by $\mathcal{Z}$, and receives reward $r\in \mathbb{R}$. A history $h_t=(ao)_{1:t-1}$ captures all the past actions and observations. The agent maintains a distribution over states given the current history as belief, $b_t(s)=Pr(s|h_t)$. After taking an action and receiving an observation, the belief is updated by $b_{t+1}(s')=\eta Pr(o|s',a)\sum_sPr(s'|s,a)b_t(s)$. Where $\eta$ is a normalizing constant. The ultimate objective of the agent is to find a policy $\pi(b_t)\in \mathcal{A}$ which maximizes the expectation of future discounted cumulative reward: 
\[
V^\pi(b_t)=\mathbb{E}\left[ \sum_{k=0}^\infty \gamma^kR(s_{t+k}, \pi(b_{t+k})|b_t  \right]
\]

Exact solutions for POMDPs are often intractable apart from some toy problems \cite{smallwoodOptimalControlPartially1973a}. Among the approximate solvers, offline solution methods \cite{pineauPointbasedValueIteration2003, kurniawatiSARSOPEfficientPointBased} act increasingly slow for larger domains. Online POMDP solvers such as POMCP \cite{silver2010monte} leverage sparse belief sampling and Monte Carlo Tree Search to scale up for large POMDPs and maintain an unweighted particle filter \cite{thrunProbabilisticRobotics2002} based belief to handle the curse of history.
%
POMCP allows us to solve the POMDP without an explicit transition and observation
model for discrete state, action, and observation spaces. It only needs a
black-box simulator that returns the next state, observation, and reward given a
state and action, $(s_{t+1}, o_{t+1}, r_{t+1}) \sim \mathcal{G}(s_t, a_t)$. In our
specific application, we adopt the POMCP solution approach while explicitly
modeling the observation likelihood $\mathcal{Z}(o \mid s, a, s')$ as a part of
$\mathcal{G}$.

\section{Related Works}
A rich body of literature is dedicated to explaining the spatial attention mechanism demonstrated by humans. The attention spotlight theory~\cite{posnerAttentionDetectionSignals} suggested that attention can be focused on a specific area of the visual field to enhance the processing of information. The zoom lens model of attention~\cite{eriksenVisualAttentionField1986} later suggested that the size of the attentional focus and the processing efficiency within the focus area can vary. In the meantime, feature integration theory~\cite{treisman1980feature} highlighted that search time for an object increases linearly with the number of distractors (background objects possessing at least one feature with the object), which is explained by the sequential nature of attention deployment. This theory was further refined with the guided search model~\cite{wolfeGuidedSearch201994,wolfe2007guided}, which suggested that the order of sequential attention deployment is determined by the saliency of the attention map and the search time is more influenced by the signal-to-noise ratio rather than the number of distractors. In addition to spatial attention, search time can be greatly improved by suppressing irrelevant information, which can be aided by negative cues~\cite{arita2012templates} and active suppression~\cite{gaspelin2015direct}. 

These studies motivated leveraging the spatial attention mechanism to formulate computational models of eye movement for information maximization~\cite{najemnik2005optimal}. The I-POMDP framework~\cite{butkoIPOMDPInfomaxModel2008} adopted a POMDP formulation for sequential eye movement modeling to maximize long-term information gathering. In another subsequent work~\cite{butko2009optimal}, the I-POMDP framework was extended to computer vision tasks through the sequential fixation of a digital fovea, which retains resolution at the center of the fixation point while downscaling toward the periphery. Notably, these works focused on retaining resolution in already acquired images and did not consider further active enhancement before acquisition, as the retained resolution may still be insufficient for critical decision-making. Additionally, the sequential fixation of the digital fovea followed an offline-learned policy to generate actions from belief, which requires extensive training and lacks interpretability.

Within the robotics community, a number of works focused on task-specific image acquisition and observation model formulation under partial observability. In \cite{zheng2021multi}, the authors used POMCP for object search in an indoor environment. They proposed a multi-object search framework from an object-oriented POMDP approach \cite{wandzelMultiObjectSearchUsing2019}. However, the work assumed a properly functioning sensor and considered that the belief of the object should increase if the object is not occluded within the FOV. In a recent work~\cite{nguyenLanguageConditionedObservationModels2023}, the authors considered a language-conditioned dynamic observation model and demonstrated improved object search performance compared to a fixed noise model in an indoor setting. However, they incorporated the detector's confidence score (high/low bucket), for modulating the observation noise and subsequently updating POMDP belief, which may provide uncalibrated results in an outdoor environment due to huge scale and appearance variation of the target object. Another recent work focused on UAV target search in an outdoor environment in a search and rescue (SAR) setting leveraging the POMCP framework~\cite{rayHumanCenteredAutonomyUAS2024}. However, they considered a distance-based observability approach where, if the agent is outside the defined distance, the instance of false positive detection is deterministically ruled out. One common similarity in these works is that they considered physical movement of the agent to observe the object within a reliable range, which is often infeasible for a UAV operating in an urban airspace, as frequent movement is often restricted. 

Conventional computer vision methods for small object detection approach this problem with slicing~\cite{akyon2022slicing}, which enhances the target object's scale through overlapping slicing of the input image and running the inference on each slice to find the target, assuming a sufficiently high-resolution image is available, or coarse-to-fine methods~\cite{huang2022ufpmp}, which detect candidate clusters with a coarse object detector and finally find the target with a fine pass of object detection. However, these methods still struggle due to resolution downscaling and can produce uncalibrated confidence scores with a number of false positives. Though there are more expensive methods to enhance prediction accuracy, such as running inference with multiple models commonly known as ensemble~\cite{solovyev2019weighted}, which are trained on a subset of the dataset~\cite{breiman1996bagging} or with random initialization~\cite{lakshminarayanan2017simple}, their application is limited by onboard computational resources, and they also require sufficient features and a specific scale to generate reliable results. In this work, we bridge this gap by considering the PTZ functionality as an alternative approach for ground target search from the cruise altitude that enables the UAV to acquire images with rich features while maintaining a fixed resolution and selectively apply model ensemble only when the belief is concentrated enough as a reasoning step between true and false positives.  

The existing works focusing on PTZ camera control~\cite{sandha2023eagle}, \cite{fang2025automatic} often focus on object tracking rather than reliable detection under sensing uncertainty. Moreover, the control policies are often learned offline assuming complete observability of the environment. These approaches are insufficient in the context of sparse ground target search from a high altitude, where variation in observation uncertainty is unavoidable based on different PTZ settings. This work further addresses these challenges by formulating camera PTZ control as a POMDP, solving it online with POMCP, and explicitly modeling observation uncertainty conditioned on the target objects' scale within the FOV.

\section{Problem Formulation} We define the POMDP by discretizing the search space with $m \times m$ grids. Grids are parameterized by top-left corner coordinates $(r,c)$ and its side length $l$ in grid units. We define a family of windows: 

\begin{align*} 
    \mathcal{W}(L) = \bigl\{(r,\, c,\, l) \;:\; l \in L;\;\; 0 \leq r,\, c \leq m - l\bigr\}, \\ |\mathcal{W}(L)| = \sum_{l \in L}(m - l + 1)^2 
\end{align*} 
where, $L \subseteq \{1, \ldots, m\}$, and  the constraint $0 \leq r,\, c \leq m - l$ ensures each window lies fully within the grid. State$(\mathcal{S})$, action$(\mathcal{A})$ and observation $(\mathcal{O})$ spaces are the instantiations of $\mathcal{W}$, with observation space additionally includes a null  $(\varnothing)$ observation.  We use the subscript $a$ for action windows $(r_a, c_a, l_a)$ and $o$ for observations $(r_o, c_o, l_o)$, and no subscript refers to the state. As the target is assumed static, the transition model reduces to the identity, $T(s' \mid s, a) = \mathbf{1}[\,s' = s\,]$ for all $s, s' \in \mathcal{S}$, $a \in \mathcal{A}$.

The observation model $\mathcal{Z}(o \mid s, a)$ specifies the probability of receiving observation $o$ given true target state $s$ and a scan action $a$. We define the scale ratio $\rho$ as a key variable governing the detector's reliability, where:
\begin{equation}
    \rho = \frac{l_a}{l}
\end{equation}
The object presence indicator $b_p\in \{0, 1 \}$ is defined as:
\begin{equation}
    b_p=\bigl[
        r_a \leq r \leq r_a + l_a - l \;\land\; c_a \leq c \leq c_a + l_a - l
    \bigr]
\end{equation}
Here, $b_p=1$ if the target lies fully within the scan window $[r_a,\, r_a+l_a) \times [c_a,\, c_a+l_a)$, otherwise $b_p=0$.
We assume, the detection reliability depends on both the presence indicator of the target $(b_p)$ and its apparent scale $(\rho)$. For each discretized scale bin $\rho_k$, we maintain Beta-Bernoulli priors for true detections and false alarms:
\[
\theta^+(\rho_k) \sim \text{Beta}(\alpha_{\text{hit},k},\beta_{\text{hit},k}), \quad
\theta^-(\rho_k) \sim \text{Beta}(\alpha_{\text{bg},k},\beta_{\text{bg},k}).
\]
The hyperparameters $\alpha_{\text{hit},k}$ and $\beta_{\text{hit},k}$ respectively encode the prior pseudo-counts of true and missed detections at scale bin $\rho_k$, while $\alpha_{\text{bg},k}$ and $\beta_{\text{bg},k}$ encode the corresponding pseudo-counts of false alarms and correctly rejected background observations. Through the prior assignment we can embed domain knowledge about the environment and the observation model being used.
We decompose $\mathcal{Z}$ by the helipad presence indicator $(b_p)$: \\

\noindent when $b_p= 1$:

\begin{equation}
    Z(o \mid s, a) =
    \begin{cases}
        \displaystyle
        \theta^+(\rho)\cdot
        \mathcal{N}\!\left(
            \begin{bmatrix} r_o \\ c_o \\ s_o \end{bmatrix};
            \begin{bmatrix} r   \\ c   \\ l   \end{bmatrix},\,
            \Sigma
        \right)
        & o \neq \varnothing \\[14pt]
        1 - \theta^+(\rho) & o = \varnothing
    \end{cases}
\end{equation}

\noindent when $b_p = 0$:

\begin{equation}
    Z(o \mid s, a) =
    \begin{cases}
        \dfrac{\theta^-(\rho)}{N_\mathrm{fp}} & o \neq \varnothing \\[10pt]
        1 - \theta^-(\rho)                    & o = \varnothing
    \end{cases}
\end{equation}

\noindent Here,  $\Sigma = \mathrm{diag}(\sigma_r^2,\, \sigma_c^2,\, \sigma_s^2)$ denotes the bounding box noise covariance and $N_\mathrm{fp} = \sum_{l_o=1}^{l_a}(l_a - l_o + 1)^2$ is the number of valid false-positive bounding box configurations inside the scan window.


We decompose the reward into in-tree step rewards and a terminal declaration reward, which is applied outside the POMCP tree when the belief-level termination policy triggers a declaration. The in-tree rewards are defined as:

\begin{equation}
    R(s, a) =
    \begin{cases}
        R_\mathrm{step} + R_\mathrm{det}-R_\mathrm{fp} & o \neq \varnothing \\
        R_\mathrm{step}                  & o = \varnothing
    \end{cases}
\end{equation}

\noindent where $R_\mathrm{step}<0$ is the per-step cost encouraging minimization of detector inferences. $R_\mathrm{det},R_\mathrm{fp}> R_\mathrm{step}$ are the detection bonus and the false positive penalty respectively, that encourages the planner to take actions that are more likely to generate true positive detections. We define the terminal reward outside the POMCP tree as:

\begin{equation}
    R_\mathrm{T}(\hat{s}_t) =
    \begin{cases}
        R_+  & \hat{s}_t = s^*  \\
        R_-  & \hat{s}_t \neq s^*
    \end{cases}
\end{equation}
Here, $R_+\gg R_-$ determines the terminal reward, $s^*$ denotes the true target state, and $\hat{s_t}$ denotes the declared state. Both $R(s,a)$ and $R_\mathrm{T}(\hat{s}_t)$ are discounted with $\gamma_{tree}$ and $\gamma_\tau$, which encourage early detection-generating actions inside the tree and reward target declaration in fewer steps. 

\section{Methodology}
Following the POMCP methodology~\cite{silver2010monte}, we approximate the belief $(b_t)$ as an unweighted particle set of $K$ hypotheses and update the belief as we receive new observations, as shown in Algorithm~\ref{alg:main_loop}. We compute the MAP (Maximum A Posteriori) estimate $(\hat{s}_t)$ and associated confidence $(\mathrm{conf}_t)$ directly from the particle histogram. Instead of running the PO-UCT~\cite{silver2010monte} on unifrom belief, we seed the simulation with a coarse confidence  full area scan $(a_0)$, and subsequent belief update, that serves the purpose of a saliency prior for our search.


\begin{algorithm}[ht]
    \caption{POMCP-PTZ: Main Loop}
    \label{alg:main_loop}
    \KwIn{Aerial image, detector}
    \KwOut{Declared target state $\hat{s}$}
    \tcp{Coarse-scan belief initialization}
    $B_0^{\mathrm{unif}} \leftarrow \{s^{(i)} \sim \mathcal{U}(\mathcal{S})\}_{i=1}^{K}$\;
    $o_0 \leftarrow \mathrm{Detector}(\mathrm{image},\, a_0)$\;
    $B_0 \leftarrow \mathrm{BeliefUpdate}(B_0^{\mathrm{unif}},\, a_0,\, o_0)$\;
    \For{$t = 1$ \KwTo $\mathrm{MAX\_STEPS}$}{
        \If{$\mathrm{conf}_t \geq \tau$}{
            \Return $\hat{s}_t$
        }
        $a_t \leftarrow \mathrm{POMCP\text{-}PLAN}(h,\, B_t,\, N_\mathrm{SIM})$\;
        \If{$\mathrm{conf}_t \geq \tau_e,a_t=MAP$}{
             $a_t=a_{ensemble}$
        }
        $o_t \leftarrow \mathrm{Detector}(\mathrm{image},\, a_t)$\;
        $B_{t+1} \leftarrow \mathrm{BeliefUpdate}(B_t,\, a_t,\, o_t,\, \hat{s}_t,\, \mathrm{conf}_t)$\;
        $h \leftarrow \mathrm{AdvanceTree}(h,\, a_t,\, o_t)$\;
    }
\end{algorithm}



Each step of $\texttt{POMCP-PLAN}$ runs $N_\mathrm{SIM}$ simulations from the current belief node, denoted as root node. Each simulation begins by sampling a state $s \sim B(h)$ from the root node, then traverses the tree using the PO-UCT~\cite{silver2010monte} method until a fixed depth. At each internal node, the UCB action selection criterion balances exploitation of high-value actions with exploration of under-visited ones:

\begin{equation}
    a^* = \arg\max_{a \in \mathcal{A}} \left[ V(ha) + 
    c\sqrt{\frac{\ln N(h)}{N(ha)}} \right]
\end{equation}

where $V(ha)$ is the current value estimate of history-action pair $(h, a)$, $N(h)$ is the visit count of node $h$, $N(ha)$ is the visit count of action $a$ at node $h$, and $c$ is the exploration constant. At leaf nodes, a rollout policy $\pi_\mathrm{rollout}$ using uniform random scan selection provides a value estimate. After each simulation, Q-values are backed up via incremental averaging:

\begin{equation}
    V(ha) \leftarrow V(ha) + \frac{R - V(ha)}{N(ha)}
\end{equation}

After executing action $a_t$ and receiving observation $o_t$ from the detector, the belief is updated via rejection sampling.

The rejection sampling procedure samples a particle $s$ from the current 
belief, passes it to the generative model $\mathcal{G}(s, a_t)$ to 
produce a simulated observation $o'$ according to the observation model $Z$, and accepts the resulting 
state $s'$ if the simulated observation matches observation $o_t$. This process repeats until $K$ particles are collected, approximating the true Bayesian posterior for the belief. When the belief is concentrated beyond a threshold, $\tau_e$, and \texttt{POMCP-PLAN} suggests a scan action covering the MAP with scale ratio $\rho=1$, we swap the default scan for an ensemble scan involving five models trained on different subsets of the dataset. This selective deployment ensures the ensemble is applied at a fixed target scale and only aids the reasoning when the belief is already concentrated. If $m$ different models among the five models provide a detection, we pass this as a detection. Otherwise, \texttt{Null} observation is recorded. This step is followed by regular belief update as described above.

\begin{figure}
    \centering
    \includegraphics[width=0.5\textwidth]{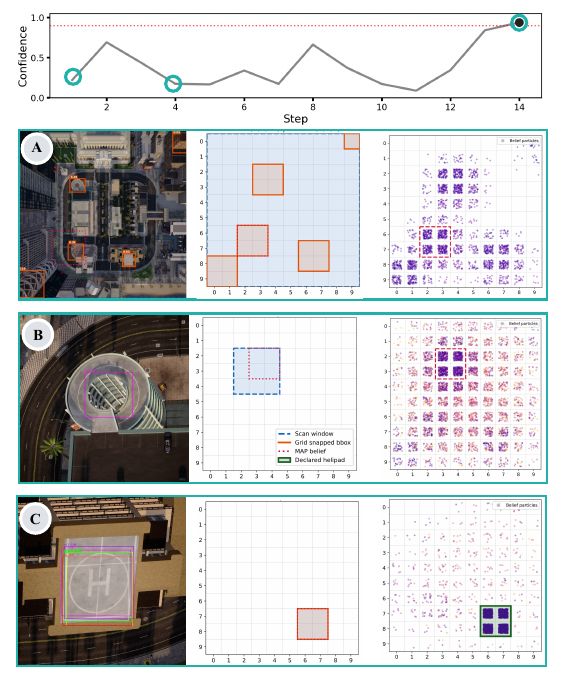}
    \caption{Evolution of belief during a full episode. The key points are marked. (A) Initial coarse scan of the area to obtain a saliency prior; this step provides true positives as well as false positive detections. The observations are snapped to the grid, and belief is updated with rejection sampling. (B) MAP concentrated on a false positive, but the ensemble returned no detection; as a result, confidence is collapsed. (C) The ensemble returned a detection at the true positive target, and belief is concentrated, resulting in target declaration. }
    \label{fig:experiment}
\end{figure}

Target declaration is handled as a belief-level termination policy applied after each 
belief update, external to the POMCP tree. The agent declares the MAP state 
$\hat{s}_t$ when belief concentration exceeds the rational threshold $\tau$. Here, $\tau$ can be tuned as a design parameter. Separating declaration from the planning tree ensures that all Q-values and UCB statistics within the tree are computed exclusively over scan actions. We illustrate key steps of an episode in Figure~\ref{fig:experiment}.

\section{Experiments and Results}

\begin{figure*}[hbt]
    \centering
    \includegraphics[width=\textwidth]{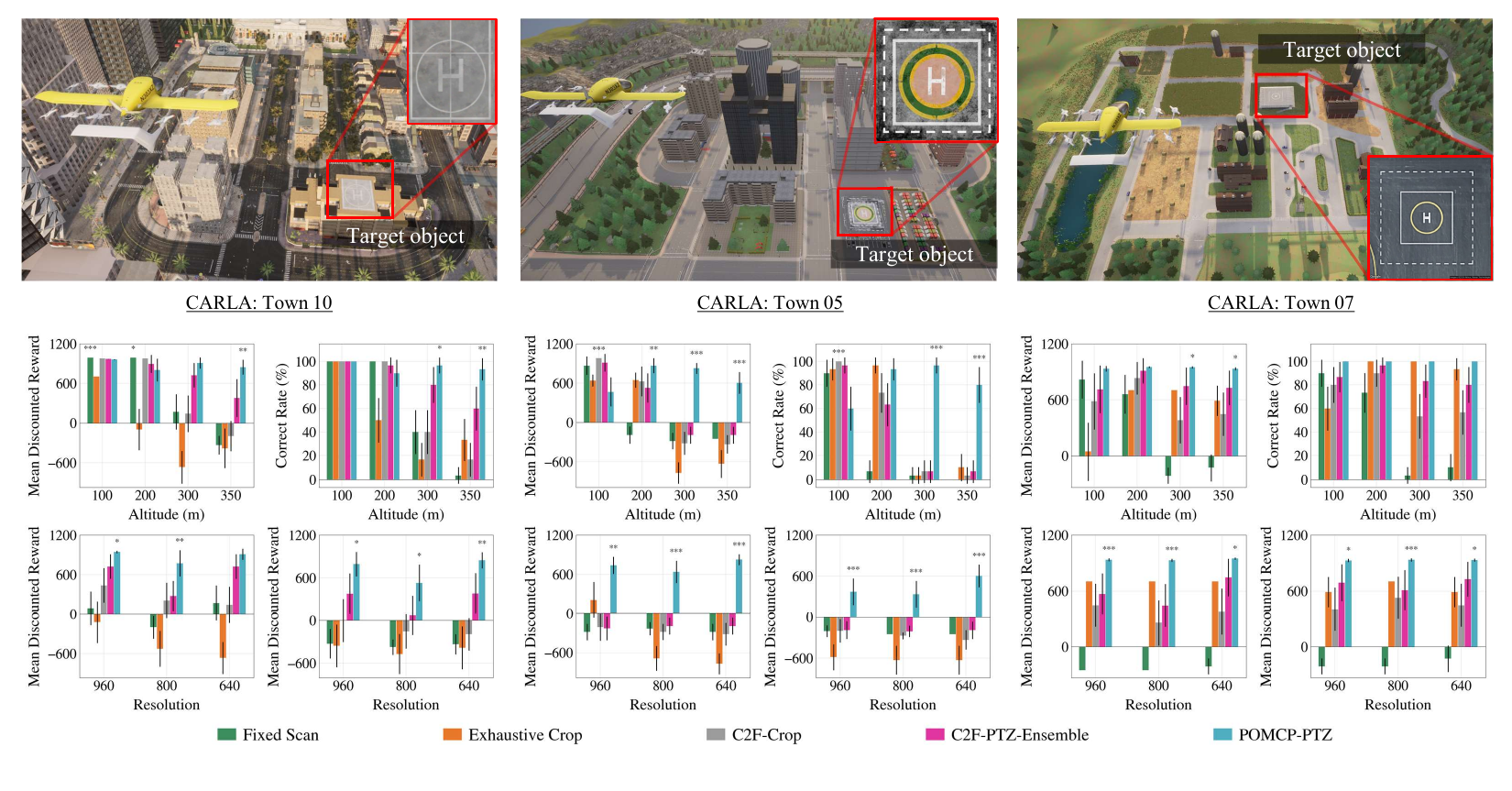}
    \caption{CARLA simulation environment and corresponding experiment results. }
    \label{fig:result}
\end{figure*}

We train six different YOLO~\cite{redmon2016} models for the experimental setup. Five models are trained on different subsets of the HelipadCat dataset~\cite{bitoun2020helipadcat} with default scale augmentation $(0.5)$, which are used for the ensemble detection. The sixth model is trianed with an optimized scale augmentation enabling the model to detect target within $(0$--$100\,m)$ altitude range following the Bayesian optimization methodology described in our previous work~\cite{rasul2025bayesian}. We first evaluate our methods within this altitude range, for which the YOLO model is already fine-tuned. Then we include higher altitudes $(200$--$350\,m)$ at a $100\,m$ increment. We keep the camera resolution fixed at $640\times 640$. We run ten trials for each method, repeat across three different seeds, and use three photorealistic environments in CARLA: Town 10, Town 05, and Town 07, as shown in Figure~\ref{fig:result}. 

We consider four different search methods as baselines. Fixed Scan method performs a single scan at full FOV and declares the object with the highest confidence score if it crosses the threshold of $0.4$. Exhaustive-Crop~\cite{akyon2022slicing} uses the slicing method at three different scales with $25\%$ overlap. A detection that is detected by most scale levels is declared as the target; in case of a tie, we declare the object with the highest mean confidence score from YOLO, with $0.4$ as the threshold value. Coarse-to-fine (C2F) methods run a coarse scan at full FOV with a low confidence threshold ($0.05$), and based on the initial detection, they enlarge the areas containing the bounding boxes to twice the initial size to produce object clusters. The clusters are then snapped to an image mosaic. Then a fine detection pass is done with $0.4$ confidence threshold and the highest confidence object is declared. We consider two different variants of C2F. The first variant, denoted as C2F-Crop, considers camera PTZ instead of slicing. The second variant, denoted as C2F-PTZ-Ensemble, considers ensemble detection at the fine pass step. 

We denote our method as POMCP-PTZ. We run each episode with $N_{SIM}=50k$ simulations with $K=11.5k$ particles. Each rollout step runs till a depth of $5$, and $MAX\_{STEPS}$ is set to $50$. We set $\alpha_f=5\%$ as a fixed particle reinvigoration rate. Our experiment results are presented in Figure~\ref{fig:result}.

At $100m$ altitude, the baseline methods perform either significantly better or as well as our method. Fixed-Scan method performs consistently, as it falls within the known altitude range within which the DNN has been fine-tuned, and it can detect the target in a single step. POMCP-based methods show comparable performance, indicating faster convergence driven by initial belief concentration from the detection received at the full-width scan. Exhaustive Crop accumulates a lower reward, discounted by the fixed high number of steps $(\approx 35)$ taken. As we increase the UAV altitude, baseline methods are no longer significantly better than POMCP-based methods, penalized by non-detection and wrong detection. At $300m$ altitude, POMCP-PTZ outperforms all other methods in Town 05 and Town 07 experiments, and provides comparable performance to c2f-PTZ-Ensemble. From this altitude, the helipad occupies only $64 \times 64$ pixels within the FOV. Therefore, Fixed Scan and Exhaustive Crop show negative reward in 2 out of 3 scenarios, penalized by repeated no-detection and wrong detection. At $350m$ altitude, the POMCP-PTZ significantly outperforms all other methods in all simulation scenarios. All other methods fail with non-detection and wrong detection as resolution is severely degraded. The C2F-Crop and C2F-PTZ-Ensemble methods, despite being allowed to access the PTZ enhancement, struggle, as even the $0.05$ confidence fails to produce an initial detection cluster due to severe resolution downscaling. On the other hand, POMCP-PTZ method chooses intermediate scans to generate initial saliency, and applies ensemble actions with tighter scans to find the target in subsequent steps. We also consider each trial as a Bernoulli trial (i.e., if the object is found correctly, we assign 100\%, and if it is not found or times out, we assign 0\%), and see the correct rate for all the methods. We observe that our method shows significantly better performance in finding the correct target in 2 out of 3 scenarios.

Though our method finds the target in fewer steps, we compare the YOLO calls to see if selective model ensemble with other methods has any severe inference overhead. Every time an ensemble action is called, we count is as 5 YOLO calls. The Fixed Scan method requires only one DNN call per episode. The Exhaustive Crop method requires 35 calls due to its fixed tiling strategy (not shown in the plot). C2F-Crop finds the target in 2 YOLO calls and C2F-PTZ-Ensemble uses 6 Yolo calls. POMCP-PTZ, though, takes significantly more YOLO calls than Fixed Scan and C2F methods; it does not take significantly more calls than the exhaustive-crop.

As slicing-based methods struggle at high altitude due to resolution downscaling, we run the next experiment to see if their performance recovers as we increase the camera resolution. The results are shown in the second row of Figure~\ref{fig:result}. We observe a steady increase in the performance of Exhaustive Crop as the resolution goes up. At higher resolution, the downscaling effect after slicing is less severe, and the DNN can detect the helipad more reliably. However, POMCP-PTZ shows only marginal improvement as it is not affected by the resolution downscaling. At higher resolution, POMCP-PTZ performance is slightly degraded (while still performing significantly better than the baselines) due to training and testing resolution mismatch.


Our experiment shows that Fixed Scan, Exhaustive Crop, and Coarse-to-Fine ensemble can be a reliable method in a known environment and altitude range. As the environment and vehicle state change, POMCP-based helipad search methods perform more reliably. Despite the fewer steps, the POMCP-based methods still have to perform simulation steps before each scan, where POMCP-PTZ has to actively pan, tilt, and zoom the camera lens. This can be undesirable if the search time is more critical than accuracy. Therefore, the choice of the search method should be made based on the specific mission requirements and the operational environment. 


\section{Conclusion}

In this work, we developed a belief-space planning algorithm for sparse target search from high altitude that incorporates observation uncertainty in the perception system with a POMDP problem formulation. Our proposed methods outperformed single-scan, traditional slicing-aided methods and coarse-to-fine methods at higher altitudes beyond the fine-tuning range and across camera resolutions.

Our proposed helipad search method is built on discretization of the state, action, and observation space, whereas in reality the state, action, and observations are continuous. This method should be further extended to POMDP solution methods, such as POMCPOW~\cite{sunbergOnlineAlgorithmsPOMDPs2018c},  that handle continuous state, action, and observation spaces requiring a higher number of simulations. The observation model considered in this work is conditioned on the object scale within the scan window, which is assumed to be conservative domain knowledge. The model itself should adapt online as it receives more feedback from the search performance. The POMCP planning method should be further optimized for speed and efficiency by incorporating different particle and simulation numbers. Physical actuation of the camera lens with PTZ operation can be a time-consuming task; therefore, both crop and PTZ can be considered within the action space for a more efficient search. The search method presented in this work is developed with a single helipad in mind. In future work, this should be extended to a multiple-helipad scenario, either by spawning multiple agents to search a single helipad each or formulating the search problem as an object-oriented POMDP~\cite{wandzelMultiObjectSearchUsing2019}. Finally, our method is developed and tested within the simulator environment only. This should be validated with real-world prototype building and testing.

\bibliographystyle{ieeetr}
\bibliography{references}

\end{document}